\pdfoutput=1

\documentclass[11pt]{article}

\usepackage[final]{acl}

\usepackage{times}
\usepackage{latexsym}

\usepackage[T1]{fontenc}

\usepackage[utf8]{inputenc}

\usepackage{microtype}

\usepackage{inconsolata}

\usepackage{graphicx}
\usepackage{amsmath}
\usepackage{amsfonts,amssymb}
\usepackage{booktabs}
\usepackage{enumerate}
\usepackage{enumitem}
\usepackage{multirow}
\usepackage{lipsum}
\usepackage{pifont}
\usepackage{color, colortbl}
\usepackage{xcolor}

\usepackage{cleveref}
\crefname{figure}{Figure}{Figures}
\Crefname{figure}{Figure}{Figures}
\crefname{table}{Table}{Tables}
\Crefname{table}{Table}{Tables}
\crefname{section}{Section}{Sections}
\Crefname{section}{Section}{Sections}
\crefname{equation}{Equation}{Equations}
\Crefname{equation}{Equation}{Equations}

\title{Predicate Debiasing in Vision-Language Models Integration for Scene Graph Generation Enhancement}

\author{Yuxuan Wang \and Xiaoyuan Liu\\
  Nanyang Technological University\\}

\newcommand{\Dsg}{\mathcal{D}_\text{sg}}

\newcommand{\Dpt}{\mathcal{D}_\text{pt}}

\newcommand{\x}{\mathbf{x}}
\newcommand{\fsg}{f_\text{sg}}
\newcommand{\fzs}{f_\text{zs}}
\newcommand{\osg}{\mathbf{o}_\text{sg}}
\newcommand{\oosg}{\mathbf{o}_\text{sg}^0}
\newcommand{\orr}{\mathbf{o}^k}
\newcommand{\orbar}{\mathbf{\hat{o}}^k}
\newcommand{\orzs}{\mathbf{o}_\text{zs}^k}
\newcommand{\orsg}{\mathbf{o}_\text{sg}^k}
\newcommand{\orzsbar}{\mathbf{\hat{o}}_\text{zs}^k}
\newcommand{\orsgbar}{\mathbf{\hat{o}}_\text{sg}^k}

\newcommand{\Ptr}{P_\text{tr}}
\newcommand{\Dtr}{\mathcal{D}_\text{tr}}
\newcommand{\Pte}{P_\text{ta}}

\newcommand{\pipt}{\mathbf{\pi_\text{pt}}}
\newcommand{\pisg}{\mathbf{\pi_\text{sg}}}

\definecolor{tabhighlight}{gray}{0.9}
\newcommand{\ensmethod}{Certainty-aware Ensemble}

\begin{document}
\maketitle
\begin{abstract}

Scene Graph Generation (SGG) provides basic language representation of visual scenes, requiring models to grasp complex and diverse semantics between objects.
This complexity and diversity in SGG leads to underrepresentation, where parts of triplet labels are rare or even unseen during training, resulting in imprecise predictions. 
To tackle this, we propose integrating the pretrained Vision-language Models to enhance representation. 
However, due to the gap between pretraining and SGG, direct inference of pretrained VLMs on SGG leads to severe bias, which stems from the imbalanced predicates distribution in the pretraining language set.
To alleviate the bias, we introduce a novel \textbf{LM Estimation} to approximate the unattainable predicates distribution. Finally, we ensemble the debiased VLMs with SGG models to enhance the representation, where we design a \textbf{certainty-aware} indicator to score each sample and dynamically adjust the ensemble weights.
Our training-free method effectively addresses the predicates bias in pretrained VLMs, enhances SGG's representation, and significantly improve the performance.

\end{abstract}

\section{Introduction}

Scene Graph Generation (SGG) is a fundamental vision-language task that has attracted much effort. 
It bridges natural languages with scene representations and serves various applications, from robotic contextual awareness to helping visually impaired people.
The key challenge in SGG is to grasp complex semantics to understand inter-object relationships in a scene.

Existing researches in SGG focus primarily on refining model architectures that are trained from scratch with datasets like Visual Genome \cite{krishna2017visual} or Open Images \cite{kuznetsova2020open}. 
However, SGG tasks inherently face another challenge of underrepresentation. Due to the inherent complexities of SGG, there exists exponential variability of triplets combined by the \textit{subject}, \textit{object}, and \textit{relation (predicate)}. It is extremely challenging for a training set to cover such diversity. As a result, a part of the test distribution is underrepresented in training, leading to poor prediction quality. In a severe case, some triplet labels that appear in the test set are unseen in training.

\begin{figure}
  \centering
  \includegraphics[width=0.475\textwidth]{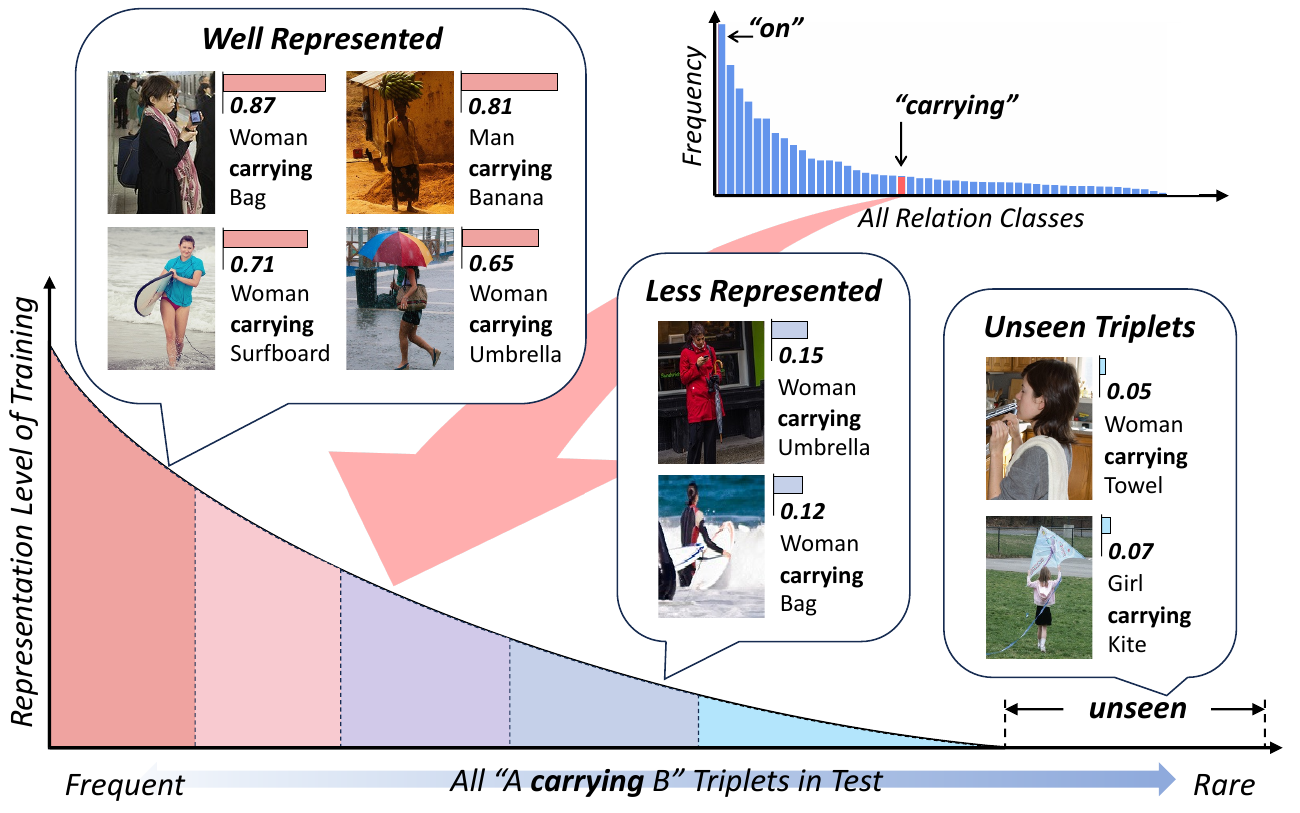}
  \caption{Illustration of the underrepresentation issue in Visual Genome. We highlight the relation class \textit{``carrying"} from the \textit{top-right} imbalanced class distribution. We present various samples with their training representation levels and confidence scores for the ground truth class, where lower scores indicate poorer prediction quality. We find that samples less represented by the training set tend to have lower-quality predictions.}
  \label{fig:intro}
\end{figure}
In Figure~\ref{fig:intro}, we highlight the relation class ``\textit{carrying}'' from Visual Genome, showing samples and their confidence scores of the ground truth class from a baseline model's predictions.
While well-represented samples score higher, the samples labeled with unseen triplets like ``woman \textit{carrying} towel" score fairly low. Furthermore, one ``woman \textit{carrying} umbrella" scores only $0.15$ due to the umbrella being closed, while its counterpart with an open umbrella scores markedly higher ($0.65$). Although the triplet is seen in training set, the closed ``umbrella'' is still short of representation.

A straightforward solution to this issue is to expand the model's knowledge by integrating advanced vision-language models (VLMs) pretrained on extensive datasets \cite{kim2021vilt,li2020oscar, li2019visualbert, qi2020imagebert,yu2022coca,radford2021learning}, using their comprehensive knowledge to compensate for underrepresented samples.
Employing the Masked Language Modeling (MLM) prompt format, such as “woman is \texttt{[MASK]} towel,” allows for direct extraction of relation predictions from the fill-in answers provided by zero-shot VLMs, which fully preserve the pretraining knowledge.
Nonetheless, this direct inference of zero-shot models on SGG introduces significant predicate bias due to disparities in data distribution and objectives between pretraining and SGG tasks.

This predicate bias originates from the imbalanced frequency of predicates in the pretraining language set, causing the VLMs to favor the predicates that are prevalent in the pretraining data. 
Unfortunately, existing debiasing methods rely on explicit training distribution, which is often unattainable for pretrained VLMs: (1) The pretraining data are often confidential. (2) Since the pretraining objectives are different with SGG, there is no direct label correspondence from pretraining to SGG.

To alleviate the predicate bias, we introduce a novel approach named \textbf{Lagrange-Multiplier Estimation} (LM Estimation) based on constrained optimization. Since there is no explicit distribution of relation labels in the pretraining data, LM Estimation seeks to estimate a surrogate distribution of SGG predicates within VLMs. 
Upon obtaining the estimated distribution, we proceed with predicates debiasing via post-hoc logits adjustment. Our LM Estimation, as demonstrated by comprehensive experiments, is proved to be exceedingly effective in mitigating the bias for zero-shot VLMs. 

Finally, we ensemble the debiased VLMs with the SGG models to address their underrepresentation issue.
We observe that some samples are better represented by the zero-shot VLM, while others align better with the SGG model. Therefore, we propose to dynamically ensemble the two models.
For each sample, we employ a \textbf{certainty-aware} indicator to score its representation level in the pretrained VLM and the SGG model, which subsequently determines the ensemble weights.
Our contributions can be summarized as follows:

\begin{itemize}[leftmargin=12pt]
\vspace{-0.2cm}
\setlength{\itemsep}{2pt}
\setlength{\parsep}{0pt}
\setlength{\parskip}{0pt}
    \item While existing methods primarily focuses on refining model architecture, we are among the pioneers in addressing the inherent underrepresentation issue in SGG using pretrained VLMs.
    \item Towards the predicates bias underlying in the pretraining language set, we propose our LM Estimation, a concise solution to estimate the unattainable words' distribution in pretraining.
    \item We introduce a plug-and-play method that dynamically ensemble the zero-shot VLMs. Needing no further training, it minimizes the computational and memory burdens. Our method effectively enhances the representation in SGG, resulting in significant performance improvement.
\end{itemize}
\section{Related Work}

\noindent
\textbf{Scene Graph Generation (SGG)} is a fundamental task for understanding the relationships between objects in images. Various of innovations \cite{tang2019learning,gu2019scene,li2021bipartite,lin2022hl,lin2020gps,lin2022ru,zheng2023prototype,xu2017scene} have been made in supervised SGG from the Visual Genome benchmark \cite{krishna2017visual}. A typical approach involves using a Faster R-CNN \cite{sun2018face} to identify image regions as objects, followed by predicting their interrelations with a specialized network that considers their attributes and spatial context. 
Existing efforts \cite{li2021bipartite,lin2022hl,lin2022ru,zheng2023prototype} mainly focus on enhancing this prediction network. For instance,  \cite{lin2022ru} introduced a regularized unrolling approach, and  \cite{zheng2023prototype} used a prototypical network for improved representation. These models specially tailored for SGG has achieved a superior performance.

\begin{figure*}[t]
  \centering
  \includegraphics[width=0.99\textwidth]{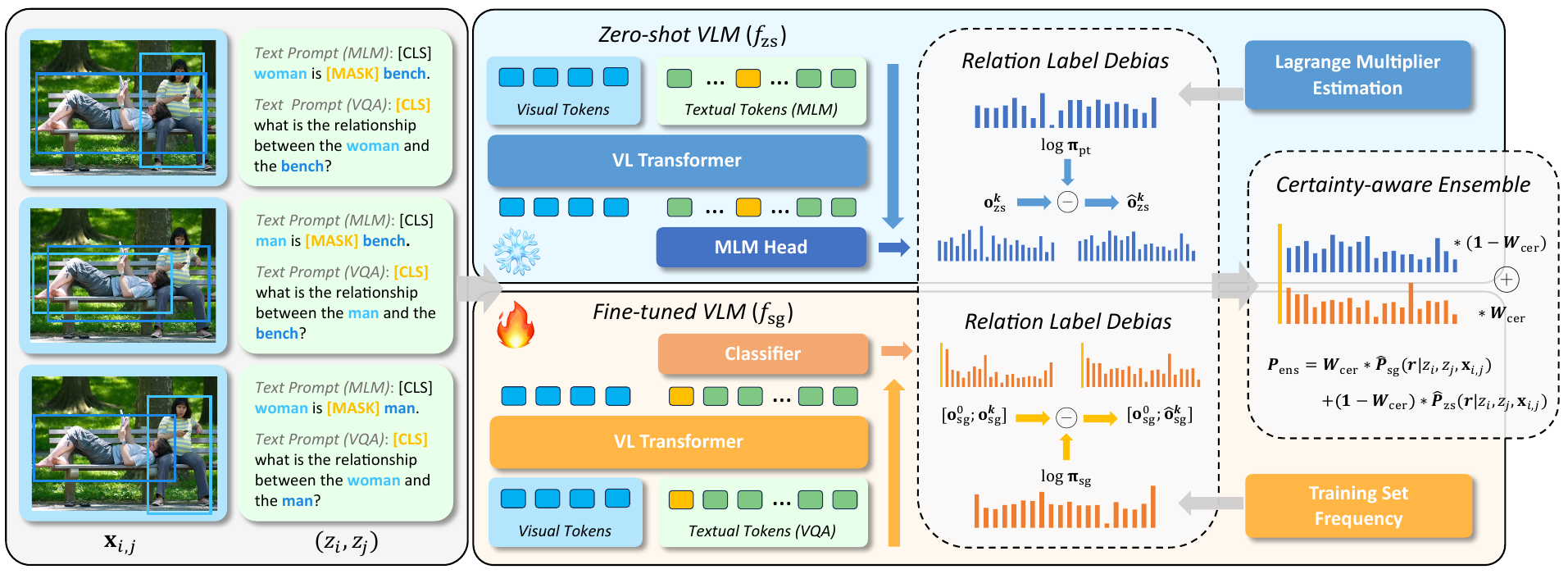}
  \caption{Illustration of our proposed architecture. \textit{left}: the visual-language inputs processed from image regions $\x_{i,j}$ and object labels $(z_i, z_j)$, either provided or predicted by Faster R-CNN detector. \textit{middle}: the fixed zero-shot VLM $\fzs$ and the trainable task-specific models $\fsg$, which we use a fine-tuned VLM as example. \textit{right}: the relation label debias process and the certainty-aware ensemble.}
  \label{fig:model}
\end{figure*}

\noindent
\textbf{Unbiased Learning} has been a long-standing challenge in classification tasks~\cite{zhu2022cross, zhu2023debiased, zhu2024generalized, zhu2023prompt}. In the specific field of SGG, started by~\cite{tang2020unbiased}, the debiasing methods~\cite{dong2022stacked,li2021bipartite,yan2020pcpl,li2022ppdl,li2022devil} seek to removing the relation label bias stemming from the imbalanced relation class distribution. These works have achieved more balanced performance across all relation classes. However, these methods rely on the interfere during training and are not feasible to the predicate bias in pre-trained VLMs. 

\noindent
\textbf{Pre-trained Vision-Language models (VLMs)} have been widely applied in diverse vision-language tasks  \cite{su2019vl, radford2021learning,kim2021vilt,li2020oscar} and have achieved substantial performance improvements with the vast knowledge base obtained during pre-training.
Recently works start to adapt the comprehensive pre-trained knowledge in VLMs to relation recognition and scene graph generation  \cite{he2022towards, gao2023compositional, li2023zero, yu2023visually, zhang2023learning, zhao2023unified}. Through prompt-tuning,  \cite{he2022towards} is the first employing VLMs to open-vocabulary scene graph generation. Then more approaches  \cite{zhang2023learning, yu2023visually, gao2023compositional} are designed towards this task. These works demonstrate the capability of VLMs on recognizing relation, inspiring us to utilize VLMs to improve the SGG representation.

\section{Methodology}

\subsection{Setup}
\label{sec:method_pre}
Given an image data $(\x, \mathcal{G})$ from a SGG dataset $\Dsg$, the image $\x$ is parsed into a scene graph $\mathcal{G}=\{\mathcal{V}, \mathcal{E}\}$, where $\mathcal{V}$ is the object set and $\mathcal{E}$ is the relation set. 
Specifically, each object $\mathbf{v} \in \mathcal{V}$ consists of a corresponding bounding box $\mathbf{b}$ and a categorical label $z$ either from annotation or predicted by a trained Faster R-CNN detector; each $\mathbf{e}_{i,j} \in \mathcal{E}$ denotes the relation for the subject-object pair $\mathbf{v}_i$ and $\mathbf{v}_j$, represented by a predicate label $y \in \mathcal{C}_e$. The predicate relation space $\mathcal{C}_e = \{0\}\cup \mathcal{C}_r $ includes one \textit{background} class $0$, indicating no relation, and $K$ \textit{non-background} relations $\mathcal{C}_r=[K]$. 
The objective is to learn a model $f$ that, given the predicted objects $z_i$ and $z_j$ for each pair with their cropped image region $\x_{i,j} =\x (\mathbf{b}_i \cup \mathbf{b}_j)$, produces logits $\mathbf{o}$ for all relations $y\in \mathcal{C}_e $, \textit{i.e.}, $\mathbf{o}=f(z_i, z_j, \x_{i,j})$.

\subsection{Method Overview}
\label{sec:overview}

As depicted in Figure~\ref{fig:model}, our framework $f$ comprising two branches: a fixed zero-shot VLM $\fzs$ and a task-specific SGG model $\fsg$ trained on $\Dsg$. Here, we employ a SGG fine-tuned VLM as $\fsg$, where we forward the image region $\x_{i,j}$ to the visual encoder and use the prompt template ``what is the relationship between the \{$z_i$\} and the \{$z_j$\}?'' as the text input. Then, a classifier head is added to the \texttt{[CLS]} token to generate logits $\osg$ of all relations $y\in \mathcal{C}_e $. Our experiments also adopt SGG models from recent works as $\fsg$.

Another zero-shot model, represented as $\fzs$, leverages pretrained knowledge to the SGG task without fine-tuning. 
By providing prompts to zero-shot VLMs in the form
``\{$z_i$\} is \texttt{[MASK]} \{$z_j$\}'', one can derive the predicted logits $\orzs$ of $K$ relation categories from the fill-in answers.
In SGG, the \textit{background} class is defined when a relation is outside $\mathcal{C}_r=[K]$. Predicting the \textit{background} relation is challenging for $\fzs$: In pretraining phase, the model has not been exposed to the specific definition of \textit{background}. Therefore, we rely solely on $\fsg$ to produce the logits of \textit{background} class:
\begin{equation}
\label{eq:two_branches}
\left\{
\begin{aligned}
    \orzs &= f_\text{zs}(z_i, z_j, \mathbf{x}_{i,j}) \in \mathbb{R}^{K}\\
    [\oosg, \orsg] &= f_\text{sg}(z_i, z_j, \mathbf{x}_{i,j}) \in \mathbb{R}^{K+1},
\end{aligned}
\right.
\end{equation}
The two branches' prediction reflect the label distribution of their training sets, leading to potential predicates bias in output logits if the target distribution differs.
To address this, we conduct predicate debiasing using our \textbf{Lagrange-Multiplier Estimation} (LM Estimation) method along with logits adjustment, generating the debiased logits $\orzsbar$ and $\orsgbar$. The details are demonstrated in Section~\ref{sec:ls problem}.

To mitigate the underrepresentation issue, we ensemble the debiased two branch to yield the final improved prediction, where we employ a \textbf{certainty-aware} indicator to dynamically adjust the ensemble weights, which is discussed in Section~\ref{sec:ensemble}.

\subsection{Predicate Debiasing}
\label{sec:ls problem}

\noindent
\textbf{Problem Definition. }
For each subject-object pair that has a \textit{non-background} relation, we denote its relation label as $r \in \mathcal{C}_r$.
Given the logits $\orr$ of $K$ \textit{non-background} relation classes, the conditional probability on the training set $\Dtr$ is computed by:
\begin{equation}\label{eq:logitsp}
    \Ptr(r|z_i, z_j, \x_{i,j}) = \text{softmax}(\orr)(r), \: r \in \mathcal{C}_r
\end{equation}
In our task, the training set $\Dtr$ can be either the SGG dataset $\Dsg$ or the pretraining dataset $\Dpt$, on which the SGG model $\fsg$ and the zero-shot model $\fzs$ are respectively trained. 

In the evaluation phase, our goal is to estimate the target test probability $\Pte$ rather than $\Ptr$.  By Bayes’ Rule, we have the following:
\begin{equation}
    P(r|z_i, z_j, \x_{i,j}) \propto P(z_i, z_j, \x_{i,j}|r) \cdot P(r)
\end{equation}
where $P \in \{\Ptr, \Pte\}$. The relation-conditional probability term $ P(z_i, z_j, \x_{i,j}|r)$ can be assumed as the same in training and testing. By changing variables and omitting the constant factor, we have:
\begin{equation}
\label{eq:ptr_pta_prop}
    \frac{\Ptr(r|z_i, z_j, \x_{i,j})}{\Ptr(r)}=\frac{\Pte(r|z_i, z_j, \x_{i,j})}{\Pte(r)}
\end{equation}
In a case where training distribution $\Ptr(r)$ not equals to the target distribution $\Pte(r)$, known as label shift, the misalignment results in the model's predicted probability $\Ptr(r|z_i, z_j, \x_{i,j})$ not equals to the actual test probability, $\Pte(r|z_i, z_j, \x_{i,j})$. 

In our framework in Figure~\ref{fig:model}, $\fzs$ is trained on $\Dpt$ and $\fsg$ on $\Dsg$, whose training label distributions $\Ptr(r)$ are $\pipt \in \mathbb{R}^{K}$ and $\pisg \in \mathbb{R}^{K}$, respectively. 
The prevalent evaluation metric, Recall, is designed to assess performance when the test label distribution $\Pte(r)$ is the same as the \textbf{training} distribution $\pisg$. 
In contrast, the mean recall metric seeks to evaluate performance in a \textbf{uniform} test distribution where $\Pte(r)=1/K$. 
The $\Ptr(r)$ and $\Pte(r)$ in each case can be summarized as follow:
\begin{equation}
\label{eq:criteria_pi}
{\Ptr(r)} = 
\begin{cases} 
{\pisg}, & \text{if } \fsg \\
{\pipt}, & \text{if } \fzs
\end{cases}, \:
{\Pte(r)} = 
\begin{cases} 
{\pisg}, & \text{training} \\
\frac{1}{K}, & \text{uniform}
\end{cases}
\end{equation}
From Equation~\ref{eq:criteria_pi}, we observe that the inequality $\Pte(r) \neq \Ptr(r)$ holds in the following scenarios:
\begin{itemize}[leftmargin=12pt]
\setlength{\itemsep}{2pt}
\setlength{\parsep}{0pt}
\setlength{\parskip}{0pt}
\item For the SGG model $\fsg$ with $\Ptr(r) = \pisg$, a label shift will be revealed when the test target is a uniform distribution evaluated by mean Recall. In this scenario, the target distribution $\Pte(r)=1/K$ diverges from the imbalanced distribution $\pisg$ in $\Dsg$ shown in \textit{top right} of Figure~\ref{fig:intro}.
\item For the zero-shot VLM $\fzs$ with $\Ptr(r) = \pipt$, the $\Pte(r) \neq \Ptr(r)$ holds in both \textbf{training} and \textbf{uniform} targets. Firstly, the label distribution $\pipt$ in the pretraining set $\Dpt$ differs from $\pisg$, resulting in $\Ptr(r)\neq\pisg$ under the training-aligned target. Secondly, the imbalanced predicates distribution in $\Dpt$ also leads to $ \Ptr(r)\neq 1/K $ under the uniform target distribution.

\end{itemize}

\noindent
\textbf{Post-hoc Logits Adjustments. }
The first case, where $ \Ptr(r)=\pisg $ but $\Pte(r)=1/K$, is a long-existing issue with many effective approaches proposed in SGG.
However, existing methods are not feasible in the second case for their debiasing in the training stage, while the pretraining stage of $\fzs$ are not accessible. 
A feasible debiasing method for already-trained models is the post-hoc logit adjustment \cite{menon2020long}. Denoting the initial prediction logits as $\orr$ and the debiased logits as $\orbar$, one can recast Equation~\ref{eq:ptr_pta_prop} into a logits form:
\begin{equation}
\label{eq:logits_adj}
    \orbar(r) = \orr(r) - \log\Ptr(r) + \log\Pte(r)
\end{equation}
It suggests that given the target label distribution, the unbiased logits $\orbar(r)$ can be obtained through a post-hoc adjustment on the initial prediction logits $\orr(r)$, following the terms' value in Equation~\ref{eq:criteria_pi}.
While $\pisg$ can be obtained simply by counting the label frequencies in $\Dsg$, $\pipt$ is the predicates distribution hidden in the pretraining stage.

\noindent
\textbf{Lagrange Multiplier Estimation. }
To accurately estimate the $\pipt$ hidden in the pretraining domain, we propose to utilize constrained optimization.
Our initial step involves collecting all samples that have \textit{non-background} relation labels $r \in \mathcal{C}_r$ from the training or validation set of $\Dsg$.
Leveraging the collected data, our optimization objective is to solve the optimal $\pipt$ that minimizes the cross-entropy loss between the adjusted logits $\orzsbar$ (following Equation~\ref{eq:criteria_pi} and~\ref{eq:logits_adj} using $\pipt$) and the ground truth relation labels $r$. 

Since the data are collected from $\Dsg$, we designate the term $\Pte(r)$ to $\pisg$ to offset the interference of its label distribution and ensure the solved $\Ptr(r)=\pipt$. 
This approach allows us to estimate  $\pipt$ by solving a constrained optimization problem, where we set the constraints to ensure the solved $\pipt$ representing a valid probability distribution:
\begin{align}
\label{eq:constrain opt}
    &\pipt = \: \underset{\pipt}{\mathrm{argmin}} \: R_{ce}(\orr - \log\pipt + \log\pisg, \: r), \notag \\
    &s.t. \: \pipt(r) \geq 0, \: \mathrm{for} \: r \in \mathcal{C}_r,\: \sum_{r \in \mathcal{C}_r}{ \pipt(r)} = 1
\end{align}
where $R_{ce}$ is the cross-entropy loss. Equation~\ref{eq:constrain opt} can be solved using the Lagrange-Multiplier method:
\begin{align}
    \pipt = \: \underset{\pipt}{\mathrm{argmin}} \:&\underset{\lambda_r \geq 0, v}{\mathrm{max}}R_{ce} - \sum_{r}{ \lambda_r\pipt(r)} \notag \\
    &+ v(1 - \sum_{r}{ \pipt(r)})
\end{align}

After obtaining $\pipt$ and $\pisg$, we can then apply the post-hoc logits adjustments for predicates debiasing following Equation~\ref{eq:criteria_pi} and~\ref{eq:logits_adj}, which produces two sets of unbiased logits from the initial prediction of $\fzs$ and $\fsg$, denoted as $\orzsbar$ and $\orsgbar$.

Upon mitigating the predicates bias inside $\fzs$, we can leverage the model to address the underrepresentation issue in $\fsg$. 
From the debiased logits $\orzsbar$ and $\orsgbar$, we compute the probabilities towards $r \in \mathcal{C}_r$, where we adopt a $\tau$-calibration outlined in \cite{kumar2022calibrated} to avoid over-confidence:
\begin{equation}
\label{eq:pte}
\left\{
\begin{aligned}
    &\hat{P}_\text{zs}(r|z_i, z_j, \x_{i,j}) = \text{softmax}(\orbar_\text{zs} / \tau)_r \\
    &\hat{P}_\text{sg}(r|z_i, z_j, \x_{i,j}) = \text{softmax}(\orbar_\text{sg} / \tau)_r
\end{aligned}
\right.
\end{equation}

\subsection{\ensmethod}
\label{sec:ensemble}
Considering that each model may better represent different samples, we compute a dynamic confidence score inspired by~\cite{hendrycks2016baseline} for each sample as its certainty in the two models, which determines the proportional weight $W_\text{cer}$ of the two models in ensemble:
\begin{equation}
\left\{
\begin{aligned}
    &\text{conf} = \underset{r \in \mathcal{C}_r}{\mathrm{max}} P(r|z_i, z_j, \x_{i,j}), P \in \{ \hat{P}_\text{zs}, \hat{P}_\text{sg} \} \\
    &W_\text{cer} \propto \: \text{sigmoid}(\text{conf}_\text{sg} - \text{conf}_\text{zs})
\end{aligned}
\right.
\end{equation}
The weights are then used to obtain the ensembled prediction on $\mathcal{C}_r$:
\begin{align}
    P_\text{ens}(r|z_i&, z_j, \x_{i,j}) = W_\text{cer}*\hat{P}_\text{sg}(r|z_i, z_j, \x_{i,j}) \notag\\ &+ (1 - W_\text{cer})*\hat{P}_\text{zs}(r|z_i, z_j, \x_{i,j})
\end{align}
Since $\fzs$ cannot predict the \textit{background} relation, we rely solely on $\fsg$ to compute the \textit{background} probability. Denoting $\osg=[\oosg, \orsg]$ as the initial logits predicted by $\fsg$ without debiasing (Equation~\ref{eq:two_branches}), the \textit{background} and \textit{non-background} probability can be calculated by softmax function:
\begin{equation}
\left\{
\begin{aligned}
    &P_\text{sg}(y\neq0|z_i, z_j, \x_{i,j}) = 1 - \text{softmax}(\osg)_0 \\
    &P_\text{sg}(y=0|z_i, z_j, \x_{i,j}) = \text{softmax}(\osg)_0
\end{aligned}
\right.
\end{equation}
Finally, the ensembled prediction on $\mathcal{C}_e$ is:
\begin{align}
\label{eq:p_decompose}
    &P_{\text{ens}}(y|z_i, z_j, \x_{i,j}) = [P_\text{sg}(y=0|z_i, z_j, \x_{i,j}), \notag\\ &P_\text{sg}(y\neq0|z_i, z_j, \x_{i,j})\cdot P_\text{ens}(r|z_i, z_j, \x_{i,j})]
\end{align}
which serves as the final representation-improved prediction of our proposed framework.

\subsection{Summary}
We integrate VLMs to mitigate the underrepresentation challenge inherent to SGG, where we propose the novel LM Estimation to approximate the unattainable pretraining distribution of predicates, $\pipt$, and conduct predicate debiasing for each model.
Unlike previous SGG methods that are optimized for one target distribution per training, our method enables seamlessly adaptation between different targets without cost, outperforming existing SGG approaches under each target distribution.

\begin{table*}[ht]
\centering
\resizebox{.99\linewidth}{!}{
\begin{tabular}{ccccccc}
\toprule
\multirow{2}{*}{Models}                  & \multicolumn{3}{c}{Predicate Classification}                       & \multicolumn{3}{c}{Scene Graph Classification}                     \\
                                         & mRecall@20           & mRecall@50           & mRecall@100          & mRecall@20           & mRecall@50           & mRecall@100          \\ \hline
VTransE\cite{zhang2017visual}                                  & 13.6                 & 17.1                 & 18.6                 & 6.6                  & 8.2                  & 8.7                  \\
SG-CogTree\cite{yu2020cogtree}                               & 22.9                 & 28.4                 & 31.0                 & 13.0                 & 15.7                 & 16.7                 \\
BGNN\cite{li2021bipartite}                                     & -                    & 30.4                 & 32.9                 & -                    & 14.3                 & 16.5                 \\
PCPL\cite{yan2020pcpl}                                     & -                    & 35.2                 & 37.8                 & -                    & 18.6                 & 19.6                 \\
Motifs-Rwt\cite{zellers2018neural}& -                    & 33.7                 & 36.1                 & -                    & 17.7                 & 19.1                 \\
Motifs-GCL\cite{dong2022stacked}                               & 30.5                 & 36.1                 & 38.2                 & 18.0                 & 20.8                 & 21.8                 \\
VCTree-TDE\cite{tang2020unbiased}                               & 18.4                 & 25.4                 & 28.7                 & 8.9                  & 12.2                 & 14.0                 \\
VCTree-GCL\cite{dong2022stacked}                               & 31.4                 & 37.1                 & 39.1                 & \textbf{19.5}        & 22.5                 & 23.5                 \\
PENET-Rwt\dag \cite{zheng2023prototype}& 31.0& 38.8                 & 40.7                 & 18.9& 22.2                 & 23.5                 \\ \hline
\multicolumn{1}{c}{Oscar ft-la} & \multicolumn{1}{c}{30.4                 } & \multicolumn{1}{c}{38.4                 } & \multicolumn{1}{c}{41.3                 } & \multicolumn{1}{c}{17.9                 } & \multicolumn{1}{c}{22.6} & \multicolumn{1}{c}{23.8                 } \\

\rowcolor{tabhighlight}
Oscar ft-la + Ours                    & 31.2(+0.8)           & 39.4(+1.0)           & 42.7(+1.4)           & 18.3(+0.4)           & 23.4(+0.8)&25.0(+1.2)           \\
 ViLT ft-la& 31.2                 & 40.5& 44.5& 17.4                 & 22.5                 &24.3\\

 \rowcolor{tabhighlight}
 ViLT ft-la + Ours                       & \textbf{32.3(+1.1)}& \textbf{42.3(+1.8)}  & \textbf{46.5(+2.0)}  & 17.9(+0.5)           & \textbf{23.5(+1.0)}&\textbf{25.5(+1.2)}  \\
 PENET-Rwt\dag& 31.4& 38.8                 & 40.7                 & 18.9& 22.2                 &23.5                 \\

\rowcolor{tabhighlight}
PENET-Rwt + Ours& 31.8(+0.4)& 39.9(+1.1)           & 42.3(+1.6)           & 19.2(+0.3)  & 23.0(+0.8)           & 24.5(+1.0)           \\ \bottomrule
\end{tabular}
}
\caption{The mean Recall results on Visual Genome comparing with state-of-the-art models and debiasing methods. The results and performance gain applying our method is below the row of corresponding baseline.
\textit{ft}: The model is fine-tuned on Visual Genome. \textit{la}: The prediction logits is debiased by logits adjustment with $\pisg$. \dag: Due to the absence of part of the results, we re-implement by ourselves.}
\label{tab:sota_mrecall}
\end{table*}

\begin{figure}
  \centering
  \includegraphics[width=0.47\textwidth]{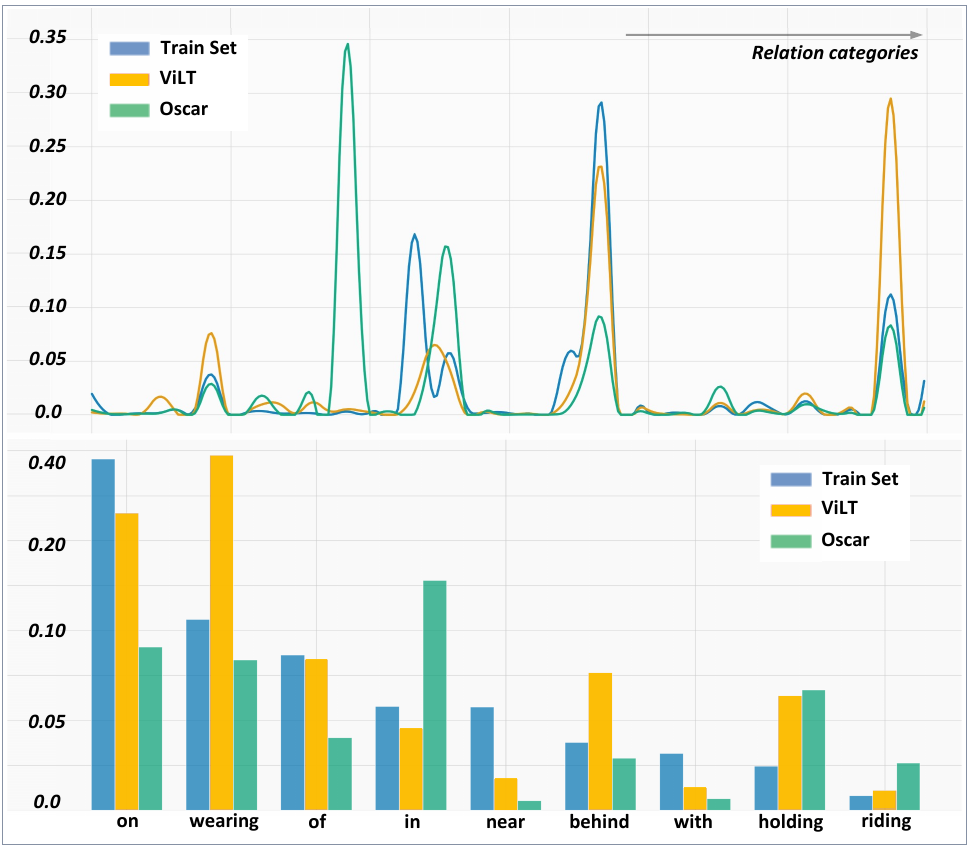}
  \caption{The relation label distributions on Visual Genome. The \textit{upper} figure illustrates the distribution across all classes, while the \textit{lower} one shows the probability distribution on some typical categories. \textit{Train Set}: The class distribution $\pisg$ in training set. \textit{ViLT} and \textit{Oscar}: The estimated distribution $\pipt$ using LM Estimation in the two pre-training stages.}
  \label{fig:label dist}
\end{figure}

\section{Experiment}

We conduct comprehensive experiments on SGG to assess our efficacy. In \cref{sec:sota}, we show our significant performance improvement through a comparative analysis with previous methods. Section~\ref{sec:estimation} provides an illustrative analysis of the predicates distribution estimated by our LM Estimation. Subsequently, \cref{sec:ablation} offers an ablation study, analysing the contribution of individual components in our design to the overall performance.

\subsection{Experiment Settings}

\noindent
\textbf{Datasets. }
The Visual Genome (VG) dataset consists of 108,077 images with average annotations of 38 objects and 22 relationships per image. 
For Visual Genome, we adopted a split with 108,077 images focusing on the most common 150 object and 50 predicate categories, allocating 70\% for training and 30\% for testing, alongside a validation set of 5,000 images extracted from the training set. 

\noindent
\textbf{Evaluation Protocol. } For the Visual Genome dataset, we focus on two key sub-tasks: Predicate Classification (PredCls) and Scene Graph Classification (SGCls). 
We skip the Scene Graph Detection (SGDet) here and provide a discussion in supplementary, considering its
substantial computational demands when employing VLMs and limited relevance to our method’s core objectives.
Our primary evaluation metrics are Recall@K and mean Recall@K (mRecall@K). 
Additionally, we propose another task of relation classification that calculates the top-1 predicate accuracy (Acc) for samples labeled with \textit{non-background} relations, where we focus on the ability of model on predicting the relation given a pair of objects in the scene.
\begin{table*}
\centering
\resizebox{.95\linewidth}{!}{
\begin{tabular}{ccccccc}
\toprule
\multirow{2}{*}{Models} & \multicolumn{3}{c}{Predicate Classification}                    & \multicolumn{3}{c}{Scene Graph Classification}                  \\
                        & Recall@20           & Recall@50           & Recall@100          & Recall@20           & Recall@50           & Recall@100          \\ \hline
KERN\cite{chen2019knowledge}                    & -                   & 65.8                & 67.6                & -                   & 36.7                & 37.4                \\
R-CAGCN\cite{yang2021probabilistic}                 & 60.2                & 66.6                & 68.3                & 35.4                & 38.3                & 39.0                \\
GPS-Net\cite{lin2020gps}                 & 60.7                & 66.9                & 68.8                & 36.1                & 39.2                & 40.1                \\
VTransE\cite{zhang2017visual}                 & 59.0                & 65.7                & 67.6                & 35.4                & 38.6                & 39.4                \\
VCTree\cite{tang2019learning}                  & 60.1                & 66.4                & 68.1                & 35.2                & 38.1                & 38.8                \\
MOTIFS\cite{zellers2018neural}                  & 59.5                & 66.0                & 67.9                & 35.8                & 39.1                & 39.9                \\
SGGNLS\cite{zhong2021learning}                  & 58.7                & 65.6                & 67.4                & 36.5                & 40.0                & 40.8                \\
RU-Net\cite{lin2022ru}                  & \textbf{61.9 }      & 68.1                & 70.1                & \textbf{38.2}       & 41.2                & 42.1                \\
PENET\dag\cite{zheng2023prototype}& 61.7                & \textbf{68.2}       & \textbf{70.1}       & 37.9                & \textbf{41.3}       & \textbf{42.3}       \\ \hline
Oscar ft& 59.1                & 65.7                & 67.6                & 36.7                & 40.3                & 41.3                \\
\rowcolor{tabhighlight}
 Oscar ft + Ours        & 60.5(+1.4)          & 67.4(+1.8)          & 69.3(+1.7)          & 37.3(+0.6)          & 41.4(+1.1)          &42.3(+1.0)          \\
 ViLT ft& 57.1                & 65.7                & 68.4                & 34.9                & 40.2                &41.8                \\
 \rowcolor{tabhighlight}
 ViLT ft + Ours          & 58.0(+0.9)          & 66.7(+1.0)          & 69.8(+1.4)          & 35.3(+0.4)          & 41.2(+1.0)          &\textbf{42.9(+1.1)} \\
PENET\dag& 61.7& 68.2& 70.1& 37.9& 41.3& 42.3\\
\rowcolor{tabhighlight}
PENET + Ours         & \textbf{62.0(+0.3)}          & \textbf{69.0(+0.8)} & \textbf{71.1(+1.0)} & \textbf{38.1(+0.2)}          & \textbf{41.8(+0.5)} & 42.9(+0.6)          \\ \bottomrule
\end{tabular}
}
\caption{The Recall results on Visual Genome dataset comparing with state-of-the-art models and debiasing methods. The results and performance gain applying our method is below the row of corresponding baseline.
\textit{ft}: The model is fine-tuned on Visual Genome. \dag: Due to the absence of part of the results, we re-implemented by ourselves.}
\label{tab:sota_recall}
\end{table*}

\noindent
\textbf{Baselines and Implementation. } 
Here we utilize two prominent zero-shot vision-language models, ViLT~\cite{kim2021vilt} and Oscar~\cite{li2020oscar}, as $\fzs$. For the task-specific branch $\fsg$, we employ three baseline models trained in SGG: 
(1) To explore the fine-tuning performance of VLMs on SGG, we fine-tune ViLT and Oscar using the PredCls training data and establish them as our first two baselines. (2) To show our methods' compatibility with existing SGG models, we undertake PENET \cite{zheng2023prototype}, a cutting-edge method with superior performance, as our third baseline.
In our ensemble strategy, we explore three combinations: "fine-tuned ViLT + zero-shot ViLT", "fine-tuned Oscar + zero-shot Oscar", and "PENET + zero-shot ViLT", where each model is debiased by our methods.
Following previous settings, an independently trained Faster R-CNN is attached to the front of each VLM model for object recognition.
During pre-training, both ViLT and Oscar employ two main paradigms: Masked Language Modeling (MLM) and Visual Question Answering (VQA). In MLM, tokens in a sentence can be replaced by \texttt{[MASK]}, with the model predicting the original token using visual and language prompts. In VQA, the model, given a question and visual input, predicts an answer via an MLP classifier using the \texttt{[CLS]} token. For our task, we use MLM for the fixed branch $\fzs$ with the prompt “$z_i$ is \texttt{[MASK]} $z_j$.” and VQA for fine-tuning $\fsg$, where we introduce a MLP with the query "\texttt{[CLS]} what is the relationship between the $z_i$ and the $z_j$?", where the embedding of \texttt{[CLS]} token is forwarded to the MLP classification head.

\subsection{Efficacy Analysis}
\label{sec:sota}

To assess the efficacy of our method, in this section, we compare our method with recent studies through a detailed result analysis on Visual Genome.
The Recall and mean Recall results are presented in Table~\ref{tab:sota_recall}, which showcases a performance comparison with a variety of cutting-edge models and debiasing methods.
We ensure to compare against previous methods under their best-performance metric. For baseline models without debiasing strategies, we compare with their superior Recall metrics and exclude their lower mean Recall performances. Similarly, for the debiased SGG models, we only focus on their mean Recall outcomes.

\noindent
\textbf{Baseline Performance. }
Our analysis begins with the three $\fsg$ baselines: fine-tuned ViLT, fine-tuned Oscar, and PENET. Specifically, for scenarios where the desired target is a uniform distribution assessed by mean Recall, we apply the post-hoc logits adjustment to the two fine-tuned baselines following Equations~\ref{eq:criteria_pi} and~\ref{eq:logits_adj}. For PENET, we implement a reweighting loss strategy (PENET-Rwt) following~\cite{zheng2023prototype} to train a debiased version tailored for the uniform target distribution, which achieved optimal performance.

Our main experiment results are presented in Table~\ref{tab:sota_mrecall} and Table~\ref{tab:sota_recall}. As shown in Table~\ref{tab:sota_recall}, without task-specific designs, the two fine-tuned VLMs fall behind the SGG models on Recall and scored 67.6 and 68.4 on R@100, while PENET takes the lead.
However, as shown in Table~\ref{tab:sota_mrecall}, when evaluated under the uniform target distribution and adjusted using simple post-hoc logits adjustment, the fine-tuned VLMs surpass all the cutting-edge debiased SGG models in mean Recall, achieving 41.3 and 44.5 of mR@100.

\begin{table*}[t]
\centering
\resizebox{.7\linewidth}{!}{
\begin{tabular}{c|cccc|cccc}
\toprule 
\multirow{2}{*}{Models} & \multicolumn{2}{c}{All mAcc} & \multicolumn{2}{c|}{All Acc} & \multicolumn{2}{c}{Unseen mAcc} & \multicolumn{2}{c}{Unseen Acc} \\
                        & Initial      & Debiased      & Initial      & Debiased      & Initial      & Debiased       & Initial      & Debiased       \\ \hline
ViLT-ft          & \multicolumn{2}{c}{46.53}    & \multicolumn{2}{c|}{68.92}   & \multicolumn{2}{c}{14.98}       & \multicolumn{2}{c}{17.72}      \\
ViLT-zs           & 21.88        & 37.42         & 57.15        & 67.09         & 8.99           & 16.92          & 18.81         & 20.93          \\
ViLT-ens          & 46.86        & 48.70         & 68.95        & 70.75         & 15.66          & 20.07          & 20.01         & 21.73          \\
\rowcolor{tabhighlight}
Ens. Gain & +0.33& +2.17& +0.03& +1.83& +0.68& +5.09& +2.29&+4.01\\ \hline
Oscar-ft         & \multicolumn{2}{c}{41.99}    & \multicolumn{2}{c|}{67.16}   & \multicolumn{2}{c}{13.85}       & \multicolumn{2}{c}{18.01}      \\
Oscar-zs          & 17.18        & 33.96         & 45.78        & 57.31         & 6.68           & 16.01          & 19.11         & 20.05          \\
Oscar-ens         & 42.02        & 44.28         & 67.77        & 69.03         & 14.83          & 19.56          & 20.97         &22.08          \\
\rowcolor{tabhighlight}
Ens. Gain & +0.03& +3.29& +0.61& +1.87& +0.98& +5.71& +2.96& +4.07\\
\bottomrule 
\end{tabular}
}
\caption{Top-1 accuracy and class-wise mean accuracy of relation classification on Visual Genome. \textit{All}: The test results for all triplets with \textit{non-background} relation labels. \textit{Unseen}: The test results for triplets that are absent from the training set. \textit{Initial}: The initial zero-shot VLMs without debiasing. \textit{Debiased}: The zero-shot VLMs after debiasing using our \textbf{LM Estimation}. \textit{ens}: Ensemble of the fine-tuned VLMs and \textit{Initial} or \textit{Debiased} zero-shot model. \textit{Ens. Gain}: the performance gain of ensemble compared to the fine-tuned model.}
\label{tab:abl_acc_tab}
\end{table*}

\noindent
\textbf{Our Improvements. }
Subsequently, we employ our certainty-aware ensemble to integrate debiased zero-shot VLMs $\fzs$ into the  $\fsg$ baselines, where each $\fzs$ is debiased by our LM Estimation. 
In Table~\ref{tab:sota_recall}, for each $\fsg$ baseline, we observed a notable performance boost after applying our methods (+1.4 / + 2.0 / + 1.6 in mR@100 and +1.7 / +1.4 / + 1.0 in R@100).
In both mRecall and Recall, our methods achieve the best performance (46.5 on mR@100 and 71.1 on R@100), while the improvement on mean Recall is particularly striking and surpasses the gains observed on Recall (+1.4/+2.0/+1.6 vs. +1.7/+1.4/+1.0). 
The results show that our methods achieve a significant improvement in each baseline, achieving the best performance compared to all existing methods.

Our results indicate the effectiveness of our methods, leading to a marked boost in performance. 
Moreover, the improvement in PENET baselines shows the adaptability of our method to existing SGG-specialized models. 
In addition, we observe that our representation improvements leads to a more significant gain in mean recall than in recall, suggesting the underrepresentation problem is more common in tail relation classes. 

\subsection{Estimated Distribution Analysis}
\label{sec:estimation}
In \cref{fig:label dist}, we depict the predicate distributions of zero-shot ViLT and Oscar solved by LM Estimation, comparing them with the distribution in VG training set. 
The \textit{upper} chart in Figure~\ref{fig:label dist} depicts the distributions across all relations, where we find that all three distributions exhibit a significant imbalance. Furthermore, we extract the distribution of typical relations in the \textit{lower} chart, where we see a substantial discrepancy among the three distributions. 
This variation affirms the two scenarios of $\Pte(r) \neq \Ptr(r)$ discussed in Section~\ref{sec:ls problem}, precluding the direct application of zero-shot VLMs without debiasing, indicating the necessity of our LM Estimation and subsequent debiasing method.

\subsection{Ablation Study}
\label{sec:ablation}
In this section, we conduct an ablation study on Visual Genome dataset.
Initially, we assess the effectiveness of our LM Estimation in addressing the predicates bias of zero-shot VLMs.
Furthermore, we evaluate the capability of our method to enhance representation by focusing on the unseen triplets, which are entirely absent during training.

To precisely evaluate the performance in relation recognition and eliminate any influence from the \textit{background} class, we require the model to perform relation classification exclusively on samples labeled with \textit{non-background} relations. Subsequently, we calculate the top-1 accuracy (Acc) and class-wise mean accuracy (mAcc) as new metrics to accurately gauge the model's effectiveness in this context. Our findings are comprehensively detailed in Table~\ref{tab:abl_acc_tab}, which details on two sample splits: one encompassing all triplets and the other exclusively focusing on unseen triplets. For each splits, we examine the performance of the two fine-tuned VLMs, $\fsg$, their initial and debiased zero-shot models, $\fzs$, and the ensemble of corresponding models.

\noindent
\textbf{Predicate Debiasing. }
In Section~\ref{sec:ls problem}, we introduce our LM Estimation method for predicate debiasing. Here, we further evaluate the efficacy of our debiasing.
We initially analysis on the relation classification accuracy of the zero-shot VLMs before and after debiasing.
As presented in Table~\ref{tab:abl_acc_tab} (the \textit{ViLT-zs} and \textit{Oscar-zs} rows), without debiasing, the accuracies of initial predictions are lower either in all triplets or unseen triplets. 
However, after debiasing through LM Estimation, there is a notable enhancement in the zero-shot performance. For unseen triplets, the debiased zero-shot VLMs even surpass the performance of their fine-tuned counterparts, suggesting our method effectively addresses the predicate bias and smoothly adapts the pretraining knowledge to the SGG task.

Furthermore, from the ensemble performance in Table~\ref{tab:abl_acc_tab} (the \textit{ViLT-ens} and \textit{Oscar-ens} rows), we notice that ensembling the initial $\fzs$ hardly improves the performance, only achieving a slight gain of +0.33/+0.03 on all triplets and +0.68/+2.29 on unseen triplets.
In contrast, ensembling the debiased $\fzs$ achieves a significantly more pronounced improvement, achieving +2.17/+1.83 gain on all triplets and +5.09/+4.01 on unseen triplets.

To keep consistent with previous settings, we present the Recall and mean Recall ablation results in Table~\ref{tab:abl_mrecall_tab}. 
We observe a substantial improvement in both mean Recall and Recall when ensembling with our debiased zero-shot VLMs (the highlighted row in each group), while directly ensembling the initial zero-shot VLMs even harm to the performance (the \textit{middle} row in each group). 
These results starkly underlines the necessity and efficacy of our LM Estimation in predicate debiasing.

\begin{table}[t]
\centering
\resizebox{.99\linewidth}{!}{
\begin{tabular}{c|ccc}
\toprule
Models& \multicolumn{1}{c}{mR@20} & \multicolumn{1}{c}{mR@50} & \multicolumn{1}{c}{mR@100} \\ \hline
ViLT-ft             & 31.2                           & 40.5                           & 44.5                            \\
ViLT-ens (Initial)         & 30.9(-0.3)& 40.5(+0.0)& 44.6(+0.1)\\
\rowcolor{tabhighlight}
ViLT-ens (Debiased)  & 32.3(+0.9)& 42.3(+1.8)& 46.5(+2.0)\\ \hline
Oscar-ft            & 30.4                           & 38.4                           & 41.3                            \\
Oscar-ens (Initial)        & 30.3(-0.1)& 38.5(+0.1)& 41.6(+0.3)\\
\rowcolor{tabhighlight}
Oscar-ens (Debiased) & 31.2(+0.8)& 39.4(+1.0)& 42.7(+1.4)\\ \midrule
Models& R@20                      & R@50                      & R@100                      \\ \hline
ViLT-ft                 & 57.1                           & 65.7                           & 68.4                            \\
ViLT-ens (Initial)             & 56.9(-0.2)& 65.7(+0.0)& 68.8(+0.4)\\
\rowcolor{tabhighlight}
ViLT-ens (Debiased)      & 58.0(+0.9)& 66.7(+1.0)& 69.8(+1.4)\\ \hline
Oscar-ft                & 59.1                           & 65.7                           & 67.6                            \\
Oscar-ens (Initial)           & 59.2(+0.1)& 65.9(+0.2)& 67.9(+0.3)\\
\rowcolor{tabhighlight}
Oscar-ens (Debiased)     & 60.5(+1.4)& 67.4(+1.7)& 69.3(+1.7)\\ \bottomrule
\end{tabular}
}
\caption{The mean Recall and Recall ablation results on Visual Genome. \textit{Initial}: The initial zero-shot VLMs without debiasing. \textit{Debiased}: The zero-shot VLMs after predicates debiasing. \textit{ens}: Ensemble of the fine-tuned VLMs and \textit{Initial} or \textit{Debiased} zero-shot model.}
\label{tab:abl_mrecall_tab}
\end{table}

\noindent
\textbf{Representation Enhancement. }
To validate the enhancement of representation, we specifically examine the samples labeled with unseen triplets. These triplets are present in the test set but absent from the training set, which is the worst tail distribution in the underrepresentation issue.

Table~\ref{tab:abl_acc_tab} reveals that, across all triplets, the accuracies of both zero-shot VLMs ($\fzs$) fall short of their fine-tuned counterparts ($\fsg$). For example, the debiased zero-shot Oscar model achieves 33.96/57.31 of mAcc/Acc, which are lower than the fine-tuned Oscar (41.99/67.16).
However, within the subset of unseen triplets, the debiased zero-shot $\fzs$ outperforms the fine-tuned $\fsg$: 
The debiased zero-shot Oscar achieves 16.01/20.05 of mAcc/Acc, outperforming the fine-tuned model (13.85/18.01).

These findings substantiate our hypothesis that zero-shot models, with their pretraining knowledge fully preserved, are better at handling underrepresented samples compared to SGG-specific models. This advantage is particularly evident in the context of unseen triplets, where comprehensive pretraining knowledge of zero-shot models confers a significant performance benefit.

Moreover, we find that the gain of ensemble is significantly higher for unseen triplets (Debiased ViLT: +5.09/+4.01, Debiased Oscar: +5.71/4.07) than for all triplets (Debiased ViLT: +2.17/+1.83, Debiased Oscar: +3.29/1.87). 
This indicates that the underrepresented samples are improved much more than the well-represented samples, receiving higher gains than average. 
Considering the proportion of unseen triplets in all triplets, we infer the overall performance gain mainly comes from the improvement on unseen triplets.
Since unseen triplets composing the worst case of underrepresentation, their performance gain can confirm our enhancement on representation.
\section{Conclusion}

In conclusion, our study has made significant strides in efficiently and effectively integrate pre-trained VLMs to SGG. 
By introducing the novel \textbf{LM Estimation}, we effectively mitigate the predicate bias inside pre-trained VLMs, allowing their comprehensive knowledge to be employed in SGG. Besides, our \textbf{certainty-aware ensemble} strategy, which ensembles the zero-shot VLMs with SGG model, effectively addresses the underrepresentation issue and demonstrates a significant improvement in SGG performance. Our work contributes to the field of SGG, suggesting potential pathways for reducing language bias of pretraining and leverage them in more complex language tasks.

\section{Limitation}

Though our methods does not require any training, comparing with original $\fsg$, our ensemble framework still adds computational cost from $\fzs$'s inference. This inference can be costly in an extreme case that one scene has too many objects to predict their relations. Besides, even after we solve the word bias inside VLMs, the final ensemble performance relies highly on the pre-training quality, which requires the $\fzs$ to be pre-trained on comprehensive data to improve SGG's representation. 
Another limitation arises from the forwarding pattern in VLM, where we adopt a pair-wise forwarding that taking a pair of objects along with their image region and text prompt. In this way, each possible object pair requires an entire forwarding of VLM. This process is rapid when the object is certainly detected. However, in the scenario of Scene Graph Detection, the large amounts of proposals can bring unavoidable time cost to our pipeline. We provide a more detailed discussion in appendix.

\bibliography{custom}

\appendix

\section{More Theoretical Justifications}

In the main paper, we introduce the post-hoc logits adjustment methods~\cite{menon2020long} for label debiasing, which is first proposed in long-tail classification. In the main paper, we skipped part of the derivation due to the limit of length. Here, we provide a detailed derivation for easier understanding.

Taking $(z_i, z_j, \x_{i,j})$ as input for a subject-object pair, the conditional probability for the relations is $P(r|z_i, z_j, \x_{i,j})$. From the Bayes' Rule, the conditional probability can be expressed as:
\begin{equation}
\label{eq:supp_1}
    P(r|z_i, z_j, \x_{i,j}) = \frac{P(z_i, z_j, \x_{i,j}|r)P(r)}{P(z_i, z_j, \x_{i,j})}
\end{equation}

We further denote the empirical probability fitted to the training set as $\Ptr$ and the target test probability as $\Pte$. We further rewrite Equation~\ref{eq:supp_1} with the two probabilities as:
\begin{align}
\label{eq:supp_2}
    \Ptr(r|z_i, z_j, \x_{i,j}) &= \frac{\Ptr(z_i, z_j, \x_{i,j}|r)\Ptr(r)}{\Ptr(z_i, z_j, \x_{i,j})}
    \\
    \Pte(r|z_i, z_j, \x_{i,j}) &= \frac{\Pte(z_i, z_j, \x_{i,j}|r)\Pte(r)}{\Pte(z_i, z_j, \x_{i,j})}
\end{align}

Then let us look into each term. Firstly, the $P(z_i, z_j, \x_{i,j})$ is irrelavant with $r$ and thus has no effect on the relation label bias. Therefore, the numerator term can be replaced by a constant $C$ and omitted in further computation. Secondly, when focusing on the label bias, according to the prevalent \textbf{label-shift hypothesis} proposed in long-tail classification, one can assume $P(z_i, z_j, \x_{i,j}|r)$ to be the same in the training and testing domains. Based on this equality, we connect the two probabilities by:
\begin{equation}
    \frac{\Ptr(r|z_i, z_j, \x_{i,j})}{\Ptr(r)} \cdot C_{\text{tr}}
    =\frac{\Pte(r|z_i, z_j, \x_{i,j})}{\Pte(r)} \cdot C_{\text{te}}
\end{equation}
Taking the logarithm form for both sides, we derive the final form of post-hoc logits adjustments~\cite{menon2020long}:
\begin{align}
\label{eq:supp_5}
    \log\Pte&(r|z_i, z_j, \x_{i,j}) = \log\Ptr(r|z_i, z_j, \x_{i,j}) \notag \\
    &- \log\Ptr(r) + \log\Pte(r) + \log\frac{C_{\text{tr}}}{C_{\text{te}}}
\end{align}
In our main paper, the last term of constant is omitted since the softmax function will naturally erase any constant term that irrelavant to $r$. 
Given the target distribution $\Pte$. From Equation~\ref{eq:supp_5}, by taking softmax operation on both sides, we can derive:
\begin{align}
\label{eq:supp_6}
    \Pte(r|z_i, z_j, \x_{i,j}\:&) = \text{softmax}(\log\Ptr(r|z_i, z_j, \x_{i,j}) \notag \\
    &- \log\Ptr(r) + \log\Pte(r))
\end{align}
After adjusting using our strategy, the final predicted label is determined by an argmax operation:
\begin{align}
\label{eq:supp_7}
    r = \underset{r \in \mathcal{C}_r}{\text{argmax}} &(\text{softmax}(\log\Ptr(r|z_i, z_j, \x_{i,j}) \notag \\
    &- \log\Ptr(r) + \log\Pte(r)))
\end{align}
Then from Equation~\ref{eq:supp_6}, we can rewrite Equation~\ref{eq:supp_7} as:
\begin{equation}
    r= \: \underset{r \in \mathcal{C}_r}{\text{argmax}}(\Pte(r|z_i, z_j, \x_{i,j}))
\end{equation}
it is called a \textbf{Bayes optimal classifier}. According to the definition of Bayes optimal classifier, on average no other classifier using the same hypothesis and prior knowledge can outperform it. Thus, when considering only label bias, our strategy is not only effective, but also optimal among all adjustments.

\section{More Experiment Analysis}

\subsection{Scene Graph Detection}
In our main paper, we skipped the SgDet sub-task, considering its substantial computational demands when employing VLMs and limited relevance to our method's core objectives. In this section, we provides a discussion and a brief corresponding experiments results.

Existing SGG models usually employs a Faster R-CNN~\cite{sun2018face} detector and fix the number of generated proposals to be 80 per image for a fair comparison. However, unlike the existing relation recognition networks that processes all pairs of proposals in an image simutaniously, the attention module in VLMs requires a one-by-one pair as input. In this case, inferencing one image requires 80$\times$80 times of forwarding.

This huge inference cost make it less practical to compare with existing methods under the current prevalent settings. However, it does not suggest using VLMs in SGG is meaningless. We strongly believe that the main concern of SGG task is to correctly recognize the relation given a pairs of objects, instead of the object detection, given the fact that the detector could be trained separately while achieving the same good performance. And by equipping with more efficient and effective detectors, the performance in Scene Graph Detection and Scene Graph Classification should be closed to Predicate Classification.

\subsection{Analysis on Tail Categories}
In this section, we conducted an additional experiment to demonstrate the performance enhancement for tail relation classes. We divided the relation categories into three splits, \textit{frequent}, \textit{medium}, and \textit{rare}, based on the frequency in the training set. Subsequently, we evaluated and reported the ensemble gain on mean Recall@100 for each split brought by our methods. We opted for mean Recall@100 as the metric due to its superior representation of rare relations and reduced susceptibility to background class interference.
Across all three baselines, we observed a substantial improvement in performance for rare relation categories, which confirms our hypothesis that the underrepresentation issue is more severe in rare relation classes.

\begin{table}[h]
\centering       
\resizebox{.8\linewidth}{!}{
\begin{tabular}{cccc}
\toprule
\multicolumn{4}{c}{Ensemble Gain on mRecall@100.}                          \\ \hline
\multicolumn{1}{c|}{Models}             & frequent & medium & rare  \\ \hline
\multicolumn{1}{c|}{ViLT ft-la + Ours}  & +0.12    & +1.78  & +4.13 \\
\multicolumn{1}{c|}{Oscar ft-la + Ours} & +0.04    & +1.04  & +3.15 \\ 
\multicolumn{1}{c|}{PENET + Ours}       & +0.06    & +1.27  & +3.49 \\ \bottomrule
\end{tabular}
}
\caption{The performance gain of mRecall@100 on PredCls sub-task achieved by our methods compared with each baseline, where the rare categories achieve significantly higher improvement.}
\label{tab:supp}
\end{table}
\section{More Details of Implementation}

This section shows more details of our implementation. In existing models designed for SGG, the object detector is attached in front of the relation recognition network and jointly trained with the objectives of SGG tasks. However, when fine-tuning VLMs on SGG tasks, this paradigm could be time-consuming and less flexible, given the higher training cost of VLM comparing with existing models. 

Therefore, we decide to take the Faster R-CNN detector out and train it separately without the main network. This implementation is proved to be effective when we take the detector out of PENET~\cite{zheng2023prototype} and train it separately with the PENET relation network. We observe that the independently trained detector achieved the same performance with that jointly trained with the PENET. Hence, all fine-tuned VLMs in this paper used a separately-trained Faster R-CNN detector. In the fine-tuning stage on Visual Genome, we employ two different paradigms for ViLT~\cite{kim2021vilt} and Oscar~\cite{li2020oscar} for a more general comparison. We freeze the ViLT backbone while training the MLP head for 50 epochs. In another way, we use an end-to-end fine-tuning for 70k steps on Oscar. We keep the fine-tuning cost comparable to the existing SGG models, which ensures its practical feasibility.

\textit{Why don't we debias on the triplets' distribution instead of the relation words distribution?} In the paper, we declare the relation words bias caused by different frequency of relation labels. And the underrepresentation issue caused by different representation level of samples. One can infer that the representation level is largely effect by the frequency of triplets. In other words, the samples of frequent triplets are usually better represented in training compared with those samples of rare triplets.
Therefore, one intuitive thinking is to debias directly on the triplets' distribution by substracting $\log P(z_i, z_j, r)$ instead of the relation words distribution $\log P(r)$. This thought is indeed the most throughly debiasing strategy. However, one need to consider that the conditional prior of $\log P(r|z_i, z_j)$ could largely help the prediction of relationship~\cite{tang2020unbiased}. For example, in natural world, the relation between a ``man" and a ``horse" is more likely to be ``man \textit{riding} horse" than ``man \textit{carrying} horse". Directly debiasing on the triplets' distribution would erase all these helpful conditional priors, resulting in a drastically drop in performance. 
\section{Other Discussions}

\noindent
\textbf{Question 1: Is our improvement from representation improvement or simply parameter increase from ensembled VLMs? }
Because of the predicates biases in pretraining data, integrating large pretrained models does not guarantee improvement. In Table 2 of the main paper, we showed that ensembling the original VLMs without debiasing cannot bring any improvements. 
Only by integrating the VLM debiased by our LM Estimation can enhancements be brought.

By integrating our debiased VLM, the underrepresentation issue is alleviated since underrepresented samples are improved much more than well-represented samples. In Table 2 in the main paper, we show that unseen triplets are improved higher than all triplets' average.
Integrating our debiased VLMs indeed brings a slight overall improvement, but most are from addressing the representation improvement. 

\noindent
\textbf{Question 2: Is it fair for us to use distinct $\Pte$ to measure Recall and mRecall and compare with existing methods? }
Unlike previous methods in SGG, our framework accepts a user-specified target distributions $\Pte$ as input.
In SGG settings, measuring both Recall and mRecall is to evaluate under two distinct test distributions, as discussed in Section 3.3 of our main paper.
For our method, using the same $\Pte$ under these two distinct distributions will input a wrong distribution $\Pte$ that is far from the actual target. This goes against our original intention.

Previous methods are measured by both metrics without any change because once trained, unless by time-costing re-training, they cannot be transferred from one target distribution $\Pte$ to another $\Pte'$. However, our method achieves this transfer instantaneously by simply $+\log{(\Pte' / \Pte)}$ to the logits. So it is fair to compare with previous methods since our transfer adds no extra time cost.

\noindent
\textbf{Question 3: Is underrepresentation issue a specific characteristic problem for SGG? }
The problem of this inadequate sample representation is a typical and specific characteristics of SGG and is far more severe than that in other related fields, like long-tailed classification in Computer Vision. 
In SGG, a sample's representation includes two objects' attributes and their high-level relationship. Due to this unique complexity, it is extremely hard for SGG datasets to adequately represent all triplets combinations. For instance, there are 375k triplets combinations in Visual Genome~\cite{krishna2017visual}, much more than the label sets of any classification dataset in Computer Vision. This inevitably leads to the majority of triplets having only a few samples in training.

\end{document}